\documentclass[runningheads]{llncs}
\usepackage{graphicx}

\usepackage{tikz}
\usepackage{comment} 
\usepackage{amsmath,amssymb} 
\usepackage{color}

\usepackage{bbm}
\usepackage{multirow}
\usepackage{graphicx}
\usepackage{xcolor,colortbl}
\usepackage{enumitem}
\usepackage{booktabs}
\newcommand{\xmark}{\ding{55}}
\usepackage{pifont}
\definecolor{amber}{rgb}{1.0, 0.75, 0.0}
\definecolor{forest}{rgb}{0, 0.75, 0}

\begin{document}
\pagestyle{headings}
\mainmatter
\def\ECCVSubNumber{2391}  

\title{UnionDet: Union-Level Detector Towards\\ Real-Time Human-Object Interaction Detection} 

\titlerunning{UnionDet}
%
\author{Bumsoo Kim\inst{1}\thanks{equal contribution, $^\dagger$corresponding author} \and
Taeho Choi\inst{1}$^{\star}$ \and
Jaewoo Kang\inst{1}$^\dagger$ \and
Hyunwoo J. Kim\inst{1}$^\dagger$}
\authorrunning{B. Kim et al.}
%
\institute{Korea University, Seoul 02841, Republic of Korea
\email{\{meliketoy,major1965,kangj,hyunwoojkim\}@korea.ac.kr}}
\maketitle

\begin{abstract}
Recent advances in deep neural networks have achieved significant progress in detecting individual objects from an image.
However, object detection is not sufficient to fully understand a visual scene.
Towards a deeper visual understanding, the interactions between objects, especially humans and objects are essential.
Most prior works have obtained this information with a bottom-up approach, where the objects are first detected and the interactions are predicted sequentially by pairing the objects.
This is a major bottleneck in HOI detection inference time.
To tackle this problem, we propose \textit{UnionDet}, a one-stage meta-architecture for HOI detection powered by a novel union-level detector that eliminates this additional inference stage by directly capturing the region of interaction.
Our one-stage detector for human-object interaction shows a significant reduction in interaction prediction time ($4\times \sim 14\times$) while outperforming state-of-the-art methods on two public datasets: V-COCO and HICO-DET.
\keywords{visual relationships, real-time detection, human-object interaction detection, object detection}
\end{abstract}
\section{Introduction}
\label{sec:introduction}
Recent advances in deep neural networks have achieved significant progress in detecting and recognizing individual objects from an image.
However, to understand a scene, we need a deeper visual understanding that transcends the level of individual object detection. 
To understand what is happening in the image, not only do we have to accurately detect individual objects, but we also have to properly predict the interactions between the detected objects.
Among the interactions, in this paper, we focus on \textit{human-object interaction (HOI) detection} that involves the localization and classification of interactions between humans and surrounding objects. HOI detection has been formally defined in \cite{gupta2015visual} as the task to detect $\langle human, verb, object \rangle$ triplets within an image.

The main challenge of HOI detection boils down to a simple question: ``\textit{How can we localize \textbf{interactions}?}''. 
When asked to localize the area of \textit{``A person rides a horse.''}, a human can naturally spot the tight area that covers both the person and the horse he/she is riding.
This is the \textit{union region} of the interacting objects that have been considered as a representation of visual relationships from previous works~\cite{lu2016visual}, and have been widely utilized in HOI detection~\cite{gao2018ican,gupta2019no,kolesnikov2019detecting,li2019transferable,wan2019pose,wang2019deep,zhou2019relation}.
Ironically, no detector in the literature has been studied to \textit{directly} capture the union region.

All the previous HOI detectors, therefore, incorporated a multi-stage and sequential pipeline that detects the individual objects first and `associate' them to obtain the union region. 
This approach is far from intuitive and it makes HOI detectors inefficient.
The sequential pipeline of object detection and interaction prediction makes end-to-end training impossible and creates a huge bottleneck in inference time for HOI detection.
In standard object detection, one-stage detectors~\cite{lin2017focal,liu2016ssd,redmon2016you,zhang2018single,zhao2019m2det} were able to speed up two-stage detectors by eliminating the second stage while yielding a competitive performance.
Yet in HOI detection, previous multi-stage models mainly focused on performance (e.g., average precision) leaving the large gap between high-performance and real-time detection unexplored.
In this work, our goal is to fill the gap between the performance and inference time of HOI detection with a fast, single-stage model.

To this end, we propose \textbf{\textit{UnionDet:}} a one-stage meta-architecture powered by a novel \textit{union-level} detector that captures the union region of human-object interaction.
Instead of associating the object detection results by feeding each object pair into a separate neural network afterward, we directly detect interacting $\langle human, object \rangle$ pairs with our novel union-level detection framework.
This eliminates the need for heavy neural network inference after object detection and enables our model to detect interactions with minimal additional time on top of existing object detectors.
Though the union-level detection sounds intuitive, detecting the union region is much more challenging than instance-level detection. 
In this paper, we study new challenges in union-level detection and address them by
 new techniques: (i) union anchor labeling, (ii) target object classification loss and (iii) union foreground focal loss.
Based on these new methods, our proposed one-stage HOI detector achieves a $4\times \sim 14\times$ speed-up in additional inference time for interaction prediction while surpassing state-of-the-art performance on two HOI detection benchmark datasets: V-COCO (\textit{Verbs in COCO}) and HICO-DET.
The main \textbf{contributions} of our paper are threefold:
\begin{itemize}
\setlength\itemsep{0.7em}
    \item We study new technical challenges in union-level detection, including bias toward human regions, inaccuracy of standard IoU-based matching and union regions containing multiple interactions and more than two objects.
    
    \item We propose a novel \textit{union-level} detector that directly detects the interaction region. We study a new set of training techniques to address the new challenges of union-level detection.
    
    \item We propose a meta-architecture \textit{\textbf{UnionDet}} equipped with our \textit{union-level} detector. It is a single-stage HOI detector achieving $4\times\sim14\times$ speed-up in interaction prediction and the \textit{state-of-the-art} performance in two public datasets.
\end{itemize}
\section{Related Work}
\label{sec:related_work}
\subsection{One-Stage Object Detection}
\label{subsec:object_detection}

One-stage object detectors based on deep neural networks have formerly been proposed for faster detection~\cite{liu2016ssd,redmon2016you,sermanet2013overfeat}.
These detectors have achieved a significant speed-up but they often come with a considerable loss of accuracy.
One known problem is the class imbalance problem.
Since one-stage object detectors densely sample anchor boxes, foreground anchor boxes are relatively much rarer than background anchor boxes (or negative samples), unlike two-stage methods that classify only a few anchor boxes after RPN.
One common technique to resolve the class imbalance is hard negative mining which samples a few hard anchor boxes for training~\cite{ren2015faster}. 
Later, RetinaNet~\cite{lin2017focal} introduced focal loss to address the issue in a fundamental way by modifying the loss function to reduce the effect of easy negatives.
Including these efforts, various techniques have been proposed to enhance one-stage object detection frameworks~\cite{zhang2018single,zhao2019m2det}.
Recently, YOLACT~\cite{bolya2019yolact} has expanded the capacity of one-stage networks to perform instance segmentation.


\subsection{Human-Object Interactions}
\label{subsec:HOI}
Human-Object Interaction (HOI) detection has been initially proposed in \cite{gupta2015visual}.
Later, human-object detectors have been improved using human body parts~\cite{fang2018pairwise}, human appearance~\cite{gkioxari2018detecting}, instance appearance~\cite{gao2018ican} and spatial relationship of human-object pairs~\cite{gao2018ican,kolesnikov2019detecting,li2019transferable}.
Especially, InteractNet~\cite{gkioxari2018detecting} extended an existing object detector by introducing an action-specific density map to localize target objects based on the appearance of a detected human.
Note that interaction detection based on visual cues from individual boxes often suffers from the lack of contextual information.
So iCAN~\cite{gao2018ican} proposed an instance-centric attention module that extracts contextual features complementary to the features from the localized objects/humans.
GPNN~\cite{qi2018learning} proposes a Graph Parsing Neural Network for HOI recognition---a general framework that explicitly represents HOI structures with graphs and automatically infers the optimal graph structures.
Deep Contextual Attention~\cite{wang2019deep} leverages contextual information by a contextual attention framework in HOI.
Recent works in HOI have also explored external knowledge to improve the performance of HOI detection.
Since the performance of HOI detection is dependent on how well we recognize the appearance of human actions, human pose information extracted from external models~\cite{cao2017realtime,chen2018cascaded,he2017mask,fang2017rmpe,li2019crowdpose} shows meaningful improvement in performance~\cite{li2019transferable,gupta2019no,wan2019pose,zhou2019relation}.
Interactiveness Knowledge has also been implemented in previous works~\cite{li2019transferable} by adding an additional inference stage where the model learns the probability of interactiveness by combining multiple HOI training datasets.
Linguistic priors and knowledge graphs are also utilized to improve HOI detection performance.
These sources are either used directly as an additional feature~\cite{peyre2019detecting,xu2019learning,gupta2017aligned} or features to cluster the objects by their functions~\cite{bansal2020detecting}.
However, all the previous methods are multi-stage detectors focusing on accuracy and they are not suitable for real-time applications.

\section{Method}
\label{sec:method}
We now introduce our method to detect human-object-interaction. To be specific, the goal is to capture $\langle human, verb, object \rangle$ triplets from an image without any external knowledge. The standard HOI detection benchmarks (e.g., V-COCO and HICO-DET) require the localization and classification of interactions. 
In this paper, we propose a one-stage HOI detector powered by our \textit{union-level} detector, which directly detects the union region of an interacting pair. 
Since standard benchmarks require the instance-level localization of humans and objects, we \textit{parallelly} combine the union-level detector and an instance-level detector, which allows more accurate instance-level localization. 
We name this meta-architecture \textbf{UnionDet} shown in Figure \ref{fig:fig_pipeline}. 
Our UnionDet is compatible with any one-stage object detectors such as SSD~\cite{liu2016ssd}, RetinaNet~\cite{lin2017feature}, and STDN~\cite{zhou2018scale}.
For a fair comparison with baseline HOI detectors, in this paper, we implement our model based on RetinaNet with ResNet50-FPN~\cite{he2016deep,lin2017feature} since it's performance is comparably similar to Faster-RCNN---the dominant backbone network in previous works on HOI in literature.

We discuss new challenges in \textit{union-level} detection and how to address them by the components in our union-level detector, which is the union branch in UnionDet in Figure \ref{fig:fig_pipeline}. We explain how to modify a standard instance-level detector in UnionDet for HOI detection and lastly, the details of training and inference are provided.

\subsection{Challenges in Union-Level Detection}
\label{subsec:union_challenges}
\begin{figure*}
    \centering
    \includegraphics[width=\textwidth]{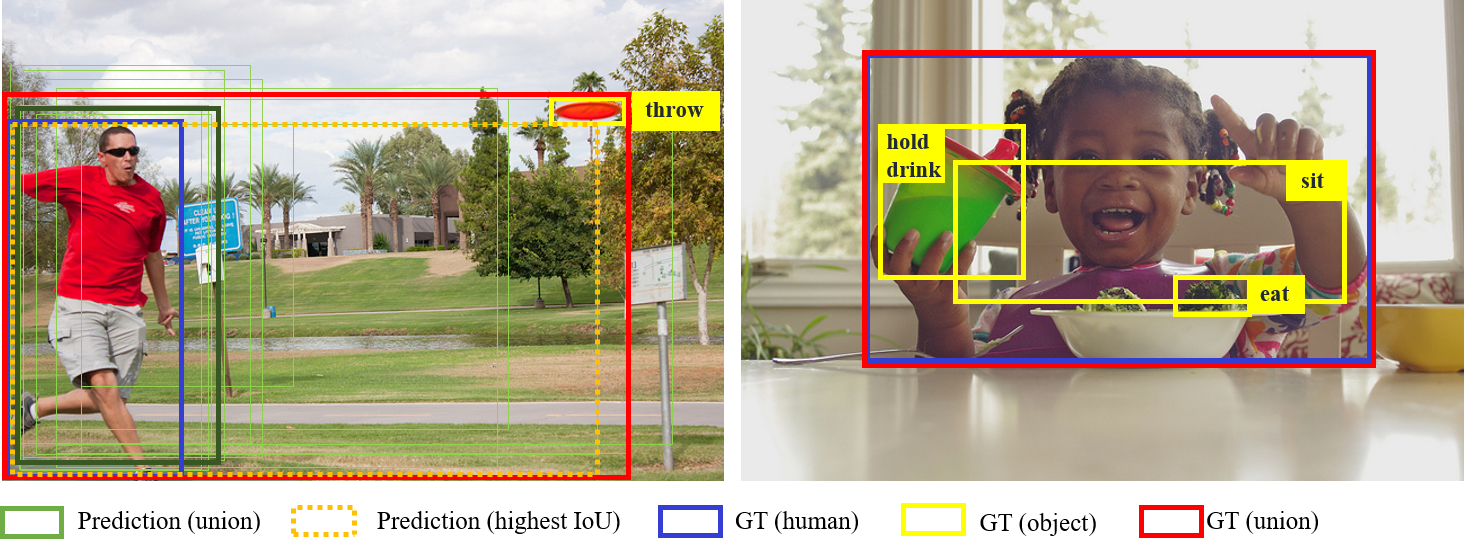}
    \caption{Technical Challenges in Union Detection.
    (Left) The box with the highest confidence for the ground-truth action 'throw' is highlighted in bold, and the region with the highest IoU with the ground-truth union region is dotted.
    As you can see, the highest confidence is biased towards the human region, and despite the high IoU, the dotted region failed to capture the target object (frisbee).
    (Right) Issues of Overlapping Union regions in \textit{One vs Many} relations.
    Even though the 3 different $\langle human, object \rangle$ pairs represent a different sets of interactions, all 3 pairs have an identical union region.
    }
    \label{fig:fig_challenges}
\end{figure*}
The union region of a pair of objects is an intuitive representation of visual relationships~\cite{lu2016visual}.
Union-level detection looks similar to instance-level detection. But standard object detectors are not directly applicable due to the following technical challenges.
\textbf{\textit{First}}, a naive union-level detection often suffers from the large bias towards human regions since every union region of HOI has a human. The left figure in Figure \ref{fig:fig_challenges} shows that union predictions (\textcolor{forest}{green} bboxes) by a vanilla detector are  densely distributed around a human.
\textbf{\textit{Second}}, the standard IoU is not an accurate metric for union bounding box matching. 
For instance, when one union region has two remote objects, a high IoU with the union region does not ensure that both human and target object is enclosed by the predicted region (see the left figure in Figure \ref{fig:fig_challenges}).
\textbf{\textit{Lastly}}, one union region (or anchor box) may contain multiple interactions and more than two objects.
These are often observed especially when a human bounding box contains multiple interacting objects.
In the following explanation of Union Branch that performs union-level detection, we show in detail how we address these issues.
\subsection{Union-level Detector: Union Branch}
\label{subsec:union_branch}
\begin{figure}
    \centering
    \includegraphics[width=\textwidth]{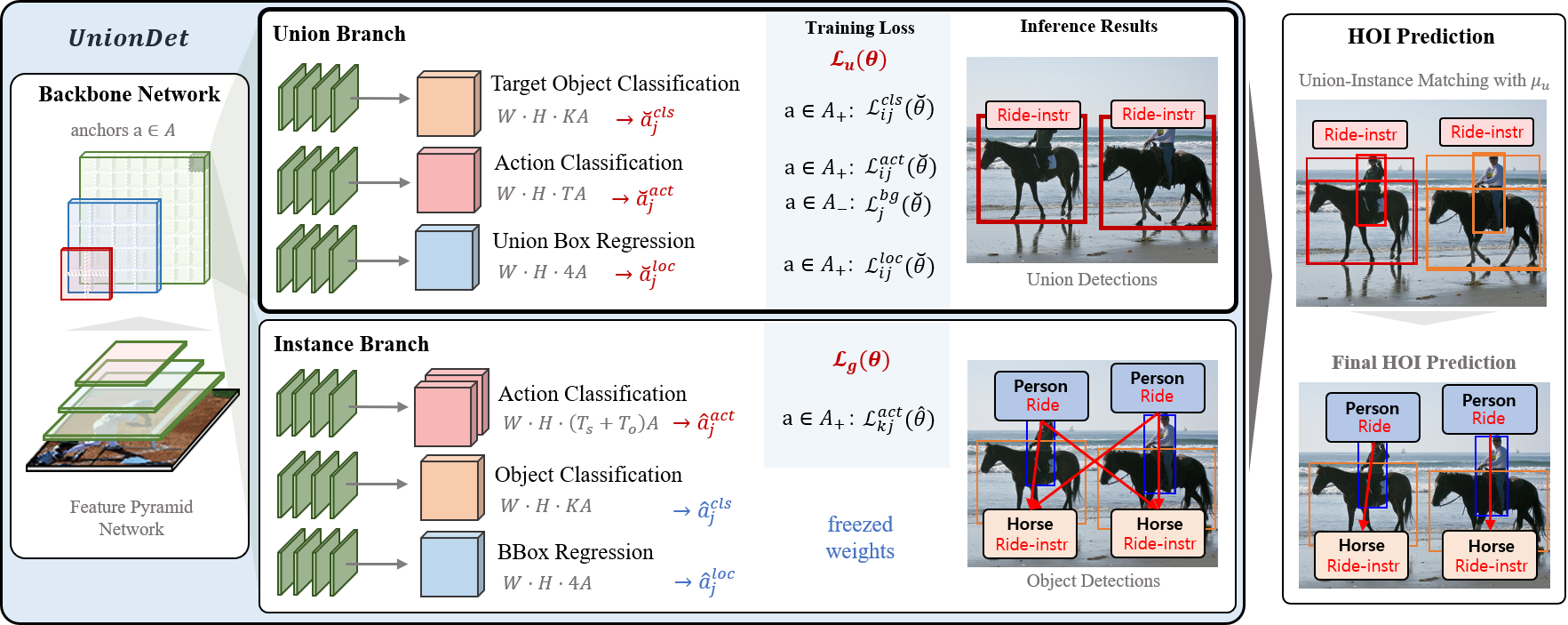}
    \caption{The overall architecture of UnionDet. Our UnionDet is generally compatible with one-stage object detectors.
    The feature pyramid obtained from the backbone network is simultaneously fed to Union Branch and Instance Branch.
    While Union Branch directly captures the region of interaction, Instance Branch performs traditional object detection and action classification for more fine-grained HOI detection results.
    }
    \label{fig:fig_pipeline}
\end{figure}

Union Branch performs union-level detection which is the essence of our proposed meta-architecture, \textit{UnionDet}.
As in Figure \ref{fig:fig_pipeline}, Union Branch consists of three sub-branches that share the backbone Feature Pyramid Network.
Out of the three sub-branches, the Action Classification sub-branch and the Union Box Regression sub-branch are the main sub-branches that contribute to the inference stage.
Action Classification sub-branch performs multi-class classification for the interactions that are related to the union region, and the Union Box Regression sub-branch performs action-agnostic bounding box regression to predict the final union region with multiple actions.
Vanilla detection results for union regions can be obtained through these two sub-branches.
However, union regions inherently accompany several technical challenges as mentioned above.
To address these challenges, Union Branch is trained by new techniques:
i) union anchor labeling
ii) target object classification loss
iii) foreground focal loss.
This provides accurate union-level detections even in various distances, see Figure \ref{fig:fig_remote}.

\subsubsection{Union Anchor Labeling.}
The standard IoU is not suitable for union-level detections. 
Especially, when objects are small and remote, an anchor box may fail to include either the subject or the object in interaction even when the ground-truth union bounding box and the anchor box has a high (e.g., 0.9) IoU.
To address this, we propose a new labeling function to match union-level labels to anchor boxes.
Union Branch detects the union regions based on the set of anchors $A$ generated from the backbone Feature Pyramid Network.
During the forward propagation of Union Branch, each anchor $a_j\in A$ obtains a multi-label action prediction $\Breve{a}^{act}_j \in \mathbb{R}^T$, target object class prediction $\Breve{a}^{cls}_j \in \mathbb{R}^K$, and a location prediction $\Breve{a}_j^{loc}\in\mathbb{R}^4$.
$T$ and $K$ denote the number of interactions and the number of target object categories, respectively.
$U_{ij}\in \{0,1\}$ indicates whether the $i_{th}$ union-level ground-truth label $\Breve{g}_i$ matches the $j_{th}$ anchor $a_j$ or not.
We propose a new anchor labeling function, which is used during training.
Let $\mathbbm{1}(\cdot)$ as an indicator function.
Given human box $\Breve{h}_i$ and object box $\Breve{o}_i$ of $i$-th ground truth union box $\Breve{g}_{i}$, $U_{ij}$ is calculated as:
\begin{equation}
\label{eq:union_anchor}
    U_{ij}=\mathbbm{1}(\mbox{IoU}(a_j,\Breve{g}_i^{loc}) > t_u)\cdot
    \mathbbm{1}\left (\frac{a_j \cap \Breve{h}_i^{loc} }{\Breve{h}_i^{loc}}>t_h \right )\cdot
    \mathbbm{1}\left (\frac{a_j \cap \Breve{o}_i^{loc}}{\Breve{o}_i^{loc}}>t_o \right ),
\end{equation}
where $t_u, t_h, t_o$  indicate thresholds for union IoU, human inclusion ratio, and object inclusion ratio. They are set to $0.5$ in our experiments.
If multiple union-level ground truths are matched, the union with the largest IoU is associated with the anchor box so that an anchor box has at most one ground truth.\newline

After labeling each anchor according to Eq.\ref{eq:union_anchor}, we can build a basic loss function to train the Union Branch.
Based on the positive anchor set $A_{+} \subseteq A$ where $\{a_j | \sum_{i}{U_{ij}}=1\}$ and the negative anchor samples $A_{-} \subseteq A$ where $\{a_j | \sum_{i} {U_{ij}}=0\}$, the loss function $\mathcal{L}_u(\Breve{\theta})$ is written as
\begin{equation}
\label{eq:vanilla}
\small
    \mathcal{L}_u(\Breve{\theta})=\sum_{a_j\in A_{+}}{
    \sum_{\Breve{g}_i\in \Breve{\mathcal{G}}}U_{ij} \left [ \mathcal{L}_{ij}^{act}(\Breve{\theta})
    + \mathcal{L}_{ij}^{loc}(\Breve{\theta})
    \right ]
    }
    + \sum_{a_j\in A_{-}}{\mathcal{L}_{j}^{bg}(\Breve{\theta})},
\end{equation}
where $\Breve{\mathcal{G}}$ denotes the ground truth union box set and $\Breve{\theta}$ denotes the Union Branch model parameters.
$\mathcal{L}^{act}_{ij}(\Breve{\theta})= FL(\Breve{a}_j^{act},\Breve{g}_i^{act},\Breve{\theta}) $,
$\mathcal{L}^{loc}_{ij}(\Breve{\theta})=smooth_{L1}(\Breve{a}_j^{loc},\Breve{g}_i^{loc},\Breve{\theta})$,
$\mathcal{L}_{j}^{bg} = FL(\Breve{a}_{j}^{act},\Vec{0},\Breve{\theta})$,
where $FL$ and $smooth_{L1}$ each denotes focal loss~\cite{lin2017focal} and Smooth L1 loss, respectively.
After training the Union Branch with Eq.\ref{eq:vanilla}, a vanilla prediction of union regions can be obtained.
However, it suffers from 1) the prediction being biased toward the human region, and 2) the noisy learning caused when multiple union regions overlap over each other.

\subsubsection{Target Object Classification Loss.}
To address the first issue where the union prediction is biased toward the human region with a vanilla union-level detector, we design a pretext task, `target object classification' from the detected union region.
This encourages the union-level detector to focus more on target objects and helps the union-level detector to capture the region that encloses the target object. 
We add the target object classification loss to Eq.\ref{eq:vanilla} and the loss function $\mathcal{L}_u$ of Union Branch is given as 
\begin{equation}
\label{eq:loss_target}
\small
    \mathcal{L}_u(\Breve{\theta})=\sum_{a_j\in A_{+}}{
    \sum_{\Breve{g}_i\in \Breve{\mathcal{G}}}U_{ij} \Big[ \mathcal{L}_{ij}^{act}(\Breve{\theta})
    + \mathcal{L}_{ij}^{loc}(\Breve{\theta})
    + \textcolor{blue}{\mathcal{L}_{ij}^{cls}(\Breve{\theta})}
    \Big]
    }
    + \sum_{a_j\in A_{-}}{\mathcal{L}_{j}^{bg}(\Breve{\theta})},
\end{equation}
where $\mathcal{L}^{cls}_{ij}(\Breve{\theta})=BCE(\Breve{a}_j^{cls},\Breve{g}_i^{cls},\Breve{\theta})$ is the  Binary Cross Entropy loss.
Though we do not use the target classification score at inference in the final HOI score function, we observed that learning to classify the target objects during training improves the union region detection as well as overall performance (see, Table \ref{tab:ablation}). 

\subsubsection{Union Foreground Focal Loss.}
Union regions often overlap over each other when a single person interacts with multiple surrounding objects.
The right subfigure in Fig.\ref{fig:fig_challenges} shows an extreme example of overlapping union regions where different interaction pairs have the exactly same union region.
In such cases where large portion of union regions overlap with each other, applying vanilla focal loss $\mathcal{L}_{ij}^{act}(\Breve{\theta})$ as in Eq.\ref{eq:vanilla} and Eq.\ref{eq:loss_target} might mistakenly give negative loss to the overlapped union actions (more detailed explanation of such cases will be dealt in our supplement).
To address this issue, we deployed a variation of focal loss where we selectively calculate losses for only positive labels for foreground regions.
This is implemented by simply multiplying $\Breve{g}_i^{act}$ to $\mathcal{L}^{act}_{ij}$, thus our final loss function is written as:
\begin{equation}
\label{eq:loss_ffl}
\small
    \mathcal{L}_u(\Breve{\theta})=\sum_{a_j\in A_{+}}{
    \sum_{\Breve{g}_i\in \Breve{\mathcal{G}}}U_{ij} \Big[ \textcolor{blue}{\Breve{g}_{i}^{act}}
    \cdot\mathcal{L}_{ij}^{act}(\Breve{\theta})
    + \mathcal{L}_{ij}^{loc}(\Breve{\theta})
    + \mathcal{L}_{ij}^{cls}(\Breve{\theta})
    \Big]
    }
    + \sum_{a_j\in A_{-}}{\mathcal{L}_{j}^{bg}(\Breve{\theta})}.
\end{equation}
\subsection{Instance-level Detector: Instance Branch}
\label{subsec:instacenet}
HOI detection benchmarks require the localization of instances in interactions. 
For more accurate instance localization, we added Instance Branch to our architecture, see Fig. \ref{fig:fig_pipeline}.
The Instance Branch parallelly performs instance-level HOI detection: object  classification, bbox regression, and action (or \textit{verb}) classification.

\subsubsection{Object Detection.}
The instance-level detector was built based on a standard anchor-based single-stage object detector that performs object classification and bounding box regression. For training, we adopt the focal loss~\cite{lin2017focal} to handle the class imbalance problem between the foreground and background anchors.
The object detector is frozen for the V-COCO dataset and fine-tuned for the HICO-DET dataset.
More discussion is available in the supplement.

\subsubsection{Action Classification.}
The instance-level detector was extended by another sub-branch for action classification.
We treat the action of subjects $T_s$ and objects $T_o$ as different types of actions.
So, the action classification sub-branch predicts $(T_s+T_o)$ action types at every anchor.
This helps to recognize the direction of interactions and can be combined with the interaction prediction from the Union Branch.
For action classification, we only calculate the loss at the positive anchor boxes where an object is located at.
This leads to more efficient loss calculation and improvement accuracy.

\subsubsection{Training Loss $\mathcal{L}_{g}$ to Learn Instance-level Actions.}
Instance Branch and Union Branch share the anchors $A$ generated from the backbone Feature Pyramid Network.
The set of instance-level ground-truth annotations for an input image is denoted as $\hat{\mathcal{G}}$.
The ground-truth label $\hat{g}_i\in \hat{\mathcal{G}}$ at anchor box $i$ consists of target class label $\hat{g}_i^{cls} \in \{ 0,1 \}^K$, multi-label action types $\hat{g}_i^{act} \in \{ 0,1 \}^{(T_s+T_o)}$ and a location $\hat{g}_i^{loc} \in \mathbb{R}^4$, i.e., $\hat{g}_i = (\hat{g}_i^{cls}, \hat{g}_i^{act}, \hat{g}_i^{loc}) \in \hat{\mathcal{G}}$.
During the forward propagation of Instance Branch, each anchor $a_j\in A$ obtains a multi-label action prediction $\hat{a}^{act}_j \in \mathbb{R}^{T_s+T_o}$ and object class prediction $\hat{a}^{cls}_j \in \mathbb{R}^K$ after sigmoid activation, and a location prediction $\hat{a}_j^{loc}=\{x,y,w,h\}$ after bounding box regression.
$I_{ij}\in \{0,1\}$ indicates whether object $\hat{g}_i$ matches anchor $a_j$ or not, i.e., $I_{ij}=\mathbbm{1}(IoU(a_j, \hat{g}_i) > t )$. We used threshold $t=0.5$ in the experiments.
The Object Classification and BBox Regression sub-branches are fixed with pre-trained weights of object detectors~\cite{lin2017focal}, and only the Action Classification sub-branch is trained.
Given parameters of Action Classification sub-branch $\hat{\theta}$, the loss for the Instance Branch $\mathcal{L}_{g}(\hat{\theta})$ will be $\mathcal{L}_{g}(\hat{\theta})=\mathcal{L}^{act}_{ij}(\hat{\theta})=BCE(\hat{a}_j^{act},\hat{g}_i^{act},\hat{\theta})$.
\subsection{Training UnionDet}
\label{training}
The two branches of UnionDet shown in Figure \ref{fig:fig_pipeline} (i.g., the Union Branch and Instance Branch) are trained jointly.
Our overall loss is the sum of the losses of both branches, $\mathcal{L}_u$ and $\mathcal{L}_g$, where $\Breve{\theta}$ is the parameters for Union Branch and $\hat{\theta}$ is the parameters for Instance Branch ($\theta = \Breve{\theta} \cup \hat{\theta}$).
The final loss becomes $\mathcal{L}(\theta)=\mathcal{L}_u(\Breve{\theta}) + \mathcal{L}_g(\hat{\theta})$.
For focal loss, we use $\alpha=0.25$, $\gamma=2.0$ as in \cite{lin2017focal}.
Our model is trained with an Adam optimizer with a learning rate of 1e-5.
\subsection{HOI Detection Inference}
\label{inference}
UnionDet at inference time parallelly performs the inference of Union Branch and Instance Branch and then seeks the highly-likely triplets using a summary score combining predictions from the subnetworks.
Instance Branch performs object detection and action classification per anchor box.
Non-maximum suppression with its object classification scores was performed.
Union Branch directly detects the union region that covers the $\langle human, verb, object \rangle$ triplet.
For Union Branch, non-maximum suppression was applied with union-level action classification scores.
Instead of applying class-wise NMS as in ordinary object detection, we treated different action classes altogether to handle multi-label predictions of union regions.

\subsubsection{Union-Instance Matching.}
As mentioned in section \ref{subsec:union_branch}, IoU is not an accurate measure for union regions, especially in the case where the target object of the interaction is remote and small.
To search for a solid union region that covers the given human box $b_h$ and object box $b_o$, we search for the union box $b_u$ with our proposed \textit{union-instance matching score} defined as 
\begin{equation}
    {\mu}_u=\frac{\text{IoU}(\lceil b_h \cup b_o \rfloor, b_u)}{2} + \frac{1}{2}\sqrt{\frac{(b_h\cap b_u)}{b_h} \cdot \frac{(b_o\cap b_u)}{b_o}},
    \label{eq:UMF}
\end{equation}
where $\frac{b_1}{b_2}$ is the ratio of the areas of two bounding boxes $b_1$ and $b_2$ and $\lceil \cdot \rfloor$ stands for the tightest bounding box that covers the area.
We use this union-instance matching score to calculate the HOI score instead of the standard IoU.

\subsubsection{HOI Score.}
The detections from Union Branch and Instance Branch are integrated. This further improves the accuracy of the final HOI detection.
Our HOI score function combines union-level action score $s^a_u$ from Union Branch with
the human category score $s_h$, human action score $s^a_h$, 
object class score $s_o$, instance-level action score $s^a_o$ from Instance Branch.
For each $\langle human, object \rangle$ pair, we first identify the best union area with the highest union-instance matching score $\mu_u$ in Eq. \eqref{eq:UMF} and then calculate the HOI score $S^a_{h,o}$ as
\begin{equation}
    S^a_{h,o}=(s_h\cdot s^a_h + s_o\cdot s_o^a)\cdot(1+{\mu}_u\cdot s^a_u).
\label{eq:hoiscore}
\end{equation}
When the action classes  do not involve target objects, or no union region is predicted, the score will be $S^a_{h,o}=s_h\cdot s^a_h$ and $S^a_{h,o}=s_h\cdot s^a_h + s_o\cdot s^a_o$, respectively.

The calculation of Eq.\eqref{eq:hoiscore} has in principle $O(n^3)$ complexity when the number of detections is $n$. 
However, our framework calculates the final triplet scores without any additional neural network inference after Union and Instance Branches. The calculation time of Eq. \eqref{eq:hoiscore} is negligible $(< 1 ms)$.
The end-to-end inference time of our model is marginally increased ($\sim 9 ms$) compared to the vanilla object detector (RetinaNet with ResNet50-FPN) thanks to the parallel architecture.
\section{Experiments}
\label{sec:experiments}
In this section, we demonstrate the effectiveness of UnionDet in HOI detection.
We first describe the two public datasets that we use as our benchmark: V-COCO and HICO-DET.
Next, we perform various qualitative and quantitative analysis to show that our union-level detector successfully addresses the proposed technical challenges and captures quality union regions, leading to a fast and accurate one-stage HOI detector.
\begin{figure}
\centering
    \includegraphics[width=\textwidth]{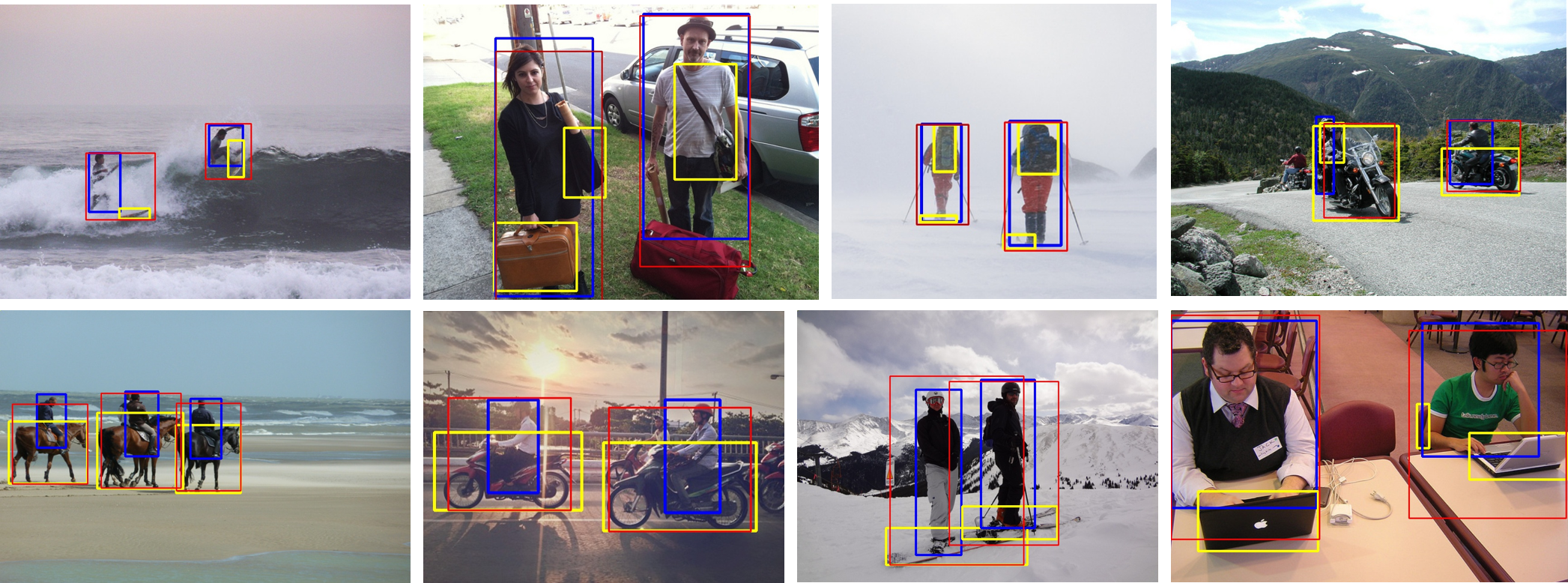}
    \caption{Union-level detections (\textcolor{red}{red}) by Union Branch successfully group 
    correct pairs of humans (\textcolor{blue}{blue}), and target objects (\textcolor{amber}{yellow}) 
    among confusing cases caused by multiple triplets with the same action and target object types in an image. Best viewed in color.}
    \label{fig:fig_disambig}
\end{figure}
\subsubsection{Datasets.}
To validate the performance of our model, we evaluate our model on two public benchmark datasets: the V-COCO (\textit{Verbs in COCO}) dataset and HICO-DET dataset.
\textbf{\textit{V-COCO}} is a subset of COCO and has 5,400 \texttt{trainval} images and 4,946 \texttt{test} images.
For V-COCO dataset, we report the $\text{AP}_{\text{role}}$ over $T=29$ interactions.
Including the four interaction types that do not involve target objects, V-COCO has $T_s=26$ active actions and $T_o=25$ passive actions.
As previous works, we exclude the interaction \textit{point} during inference time, because only 31 instances appear in the test set.
\textbf{\textit{HICO-DET}}~\cite{chao2018learning} is a subset of HICO dataset and has more than 150K annotated instances of human-object pairs in 47,051 images (37,536 training and 9,515 testing) and is annotated with 600 $\langle verb, object \rangle$ interaction types.
There are 80 unique object types, identical to the COCO object categories, and $T=117$ unique verbs.
In the HICO-DET dataset, we separate the 117 action classes into ${a_s, a_o}$, thus leading into a total action number of $T_s+T_o=234$.
For HICO-DET dataset, we follow the previous settings and report the mAP over three different category sets: (1) all 600 HOI categories in HICO (Full), (2) 138 HOI categories with less than 10 training instances (Rare), and (3) 462 HOI categories with 10 or more training instances (Non-Rare).
\subsubsection{Union-level detection.}
Our union-level detector (Union Branch in UnionDet) directly detects union regions of HOI, see 
Fig.~\ref{fig:fig_disambig}.
Interestingly, the union-level detections are useful to disambiguate the confusing pairs with the 
same action and target object types (e.g., horse, or motorcycle) in an image.
For example, when multiple people \textit{ride} the same target objects as in Fig.~\ref{fig:fig_disambig},  instance-level appearances are not sufficient to associate the correct pairs.
Union-level detections successfully group them using the context in the union-region.

\begin{figure}
\centering
    \includegraphics[width=\textwidth]{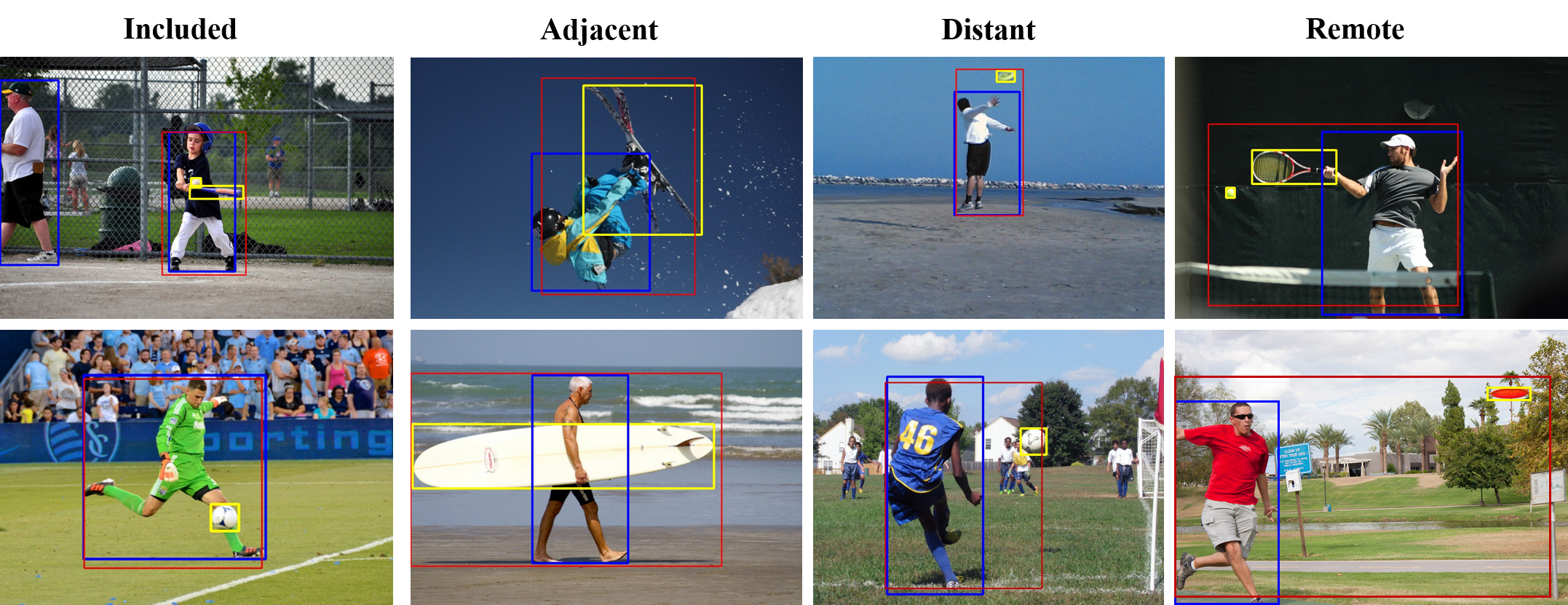}
    \caption{Our union-level detector (Union Branch) successfully detects the union bounding boxes (\textcolor{red}{red}) in various distances: the interactions with included, adjacent, distant and remote target objects.
    Also, instance-level detections of human (\textcolor{blue}{blue}), and target objects (\textcolor{amber}{yellow}) by Instance Branch are visualized. Best viewed in color.
    }
    \label{fig:fig_remote}
\end{figure}
\subsubsection{Interactions in various distances.}
We discussed in Sec.~\ref{subsec:union_challenges} that a vanilla object detector is not directly applicable to union-level detection due to the bias toward human regions. This bias gets severer especially when a human interacts with remote target objects.
Fig. \ref{fig:fig_remote} shows that the bias is addressed by our pre-text task `Target Object Classification' and UnionDet is able to detect target objects for various distances.
We show four cases: included ($b_h \supset b_o$), adjacent ($\text{IoU}(b_h,b_o) > 0$), distant ($\text{IoU}(b_h,b_o) = 0$) and remote ($\text{IoU}(b_h,b_o) = 0$ and large distance), where $b_h$, and $b_o$ are human and object bounding boxes.
Especially the fourth column in Fig. \ref{fig:fig_remote} shows that UnionDet successfully captures the remote relation with small remote target objects (e.g., tennis ball and frisbee).
Our ablation study in Table. \ref{tab:ablation} provides that the `Target Object Classification' improves HOI detection.
Qualitative results of a vanilla union-level detector without the Target Object Classification sub-branch are provided in the supplement.

\begin{figure}
    \centering
    \includegraphics[width=\textwidth]{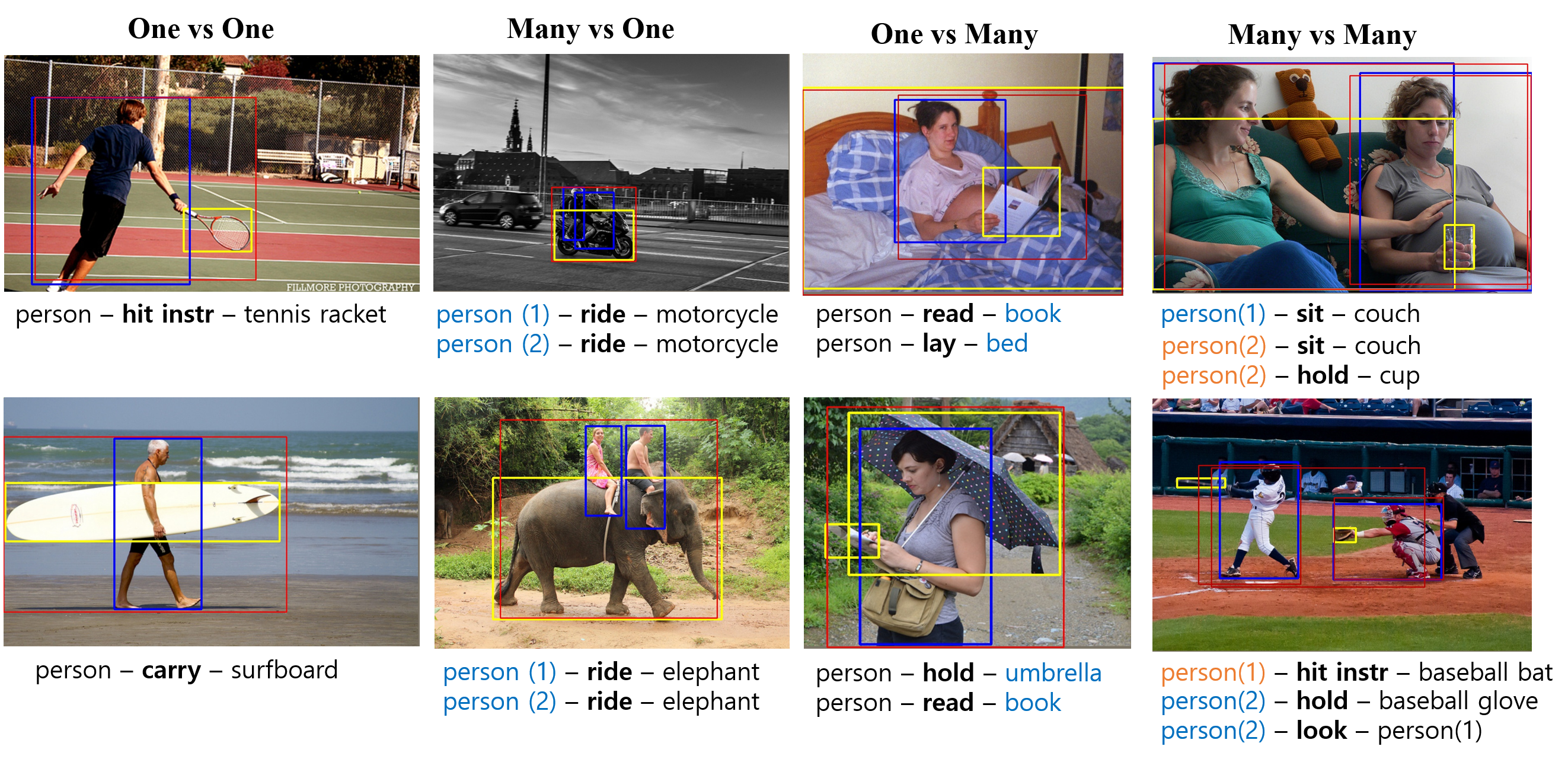}
    \caption{The final HOI detection by our model combining the predictions of both branches of UnionDet. The columns from the left to the right show one-to-one, many-to-one (multiple persons interacting with a single object), one-to-many (one person interacting with multiple objects), and many-to-many (multiple persons interacting with multiple objects) relationships respectively.
    }
    \label{fig:fig_unions}
\end{figure}
\subsubsection{HOI detection results.}
In Figure \ref{fig:fig_unions}, we highlight the detected humans and objects by object detection with the blue and yellow boxes and the union region predicted by UnionDet with red boxes.
The detected human-object interactions are visualized and given a pair of objects, the $\langle human, verb, object \rangle$ triplet with the highest HOI score is listed below each image.
Note that our model can detect various types of interactions including one-to-one, many-to-one (multiple persons interacting with a single object), one-to-many (one person interacting with multiple objects), and many-to-many (multiple persons interacting with multiple objects) relationships.
\setlength{\tabcolsep}{4pt}
\begin{table}[h!]
\caption{Comparison of performance and additional inference time on V-COCO test set. $\cdot_{\#1}$, $\cdot_{\#2}$ each refers to the performance with Scenario\#1 and Scenario\#2.
While achieving $4\times\sim 14\times$ speed-up in additional inference time for interaction prediction, our model surpasses all state-of-the-art performances in both Scenario\#1 and Scenario\#2.
Our model also achieves competitive performance to those that deploy external knowledge or features.
Note that our model does not use any external knowledge.}
\label{tab:vcoco}
  \centering
  \small
  \begin{tabular}{l|c|c|l|c}
    \toprule
    Method & Feature backbone & External Resources & $\mbox{AP}_{\mbox{role}}$ & $t$(ms) \\ \midrule\hline
    \multicolumn{5}{l}{\textit{Models with \textbf{external features}}} \\ \hline
    Verb Embedding~\cite{xu2019learning} & ResNet50 & GloVe~\cite{pennington2014glove}, VRD~\cite{lu2016visual} & {\color{gray}45.9$_{\#1}$} &  \\
    R$\text{P}_{\text{D}}\text{C}_{\text{D}}$~\cite{li2019transferable} & ResNet50 & Pose~\cite{fang2017rmpe,li2019crowdpose} & {\color{gray}47.8$_{\#1}$} &  \\
    RPNN~\cite{zhou2019relation} & ResNet50 & Keypoint~\cite{he2017mask} & {\color{gray}47.5$_{\#2}$} &  \\
    PMFNet~\cite{wan2019pose} & ResNet50-FPN & Pose~\cite{chen2018cascaded} & {\color{gray}52.0}\ &  \\ \midrule\hline
    \rowcolor[gray]{0.85}\multicolumn{5}{l}{\textit{\textbf{Models with original comparison}}} \\ \hline
    VSRL~\cite{gupta2015visual} & ResNet50-FPN & \xmark & 31.8\ & - \\
    InteractNet~\cite{gkioxari2018detecting} & ResNet50-FPN & \xmark & 40.0$_{\#2}$ & 55 \\
    BAR-CNN~\cite{kolesnikov2019detecting} & ResNet50-FPN & \xmark & 43.6\ & 125 \\
    GPNN~\cite{qi2018learning} & \textbf{ResNet152} & \xmark & 44.0\ & 40 \\
    iCAN~\cite{gao2018ican} & ResNet50 & \xmark & 44.7$_{\#1}$ & 75 \\
    TIN (R$\text{C}_{\text{D}})$~\cite{li2019transferable} & ResNet50 & \xmark & 43.2$_{\#1}$ & 70 \\
    DCA~\cite{wang2019deep} & ResNet50 & \xmark & 47.3\ & 130 \\ \midrule
    \multirow{2}{*}{\textit{\textbf{UnionDet (Ours)}}} & \multirow{2}{*}{ResNet50-FPN} & \multirow{2}{*}{\xmark} & \textbf{47.5$_{\#1}$} & \multirow{2}{*}{\textbf{9.06}} \\
     & & & \textbf{56.2$_{\#2}$} &  \\
    \bottomrule
  \end{tabular}
\end{table}
\setlength{\tabcolsep}{4pt}
\begin{table}[h!]
\caption{Performance and additional inference time comparison in HICO-DET.
Models with $\dagger$ used a heavier feature extraction backbone (i.e., ResNet152-FPN).
Further experimental settings are discussed in detail in our supplement.
Our model shows the fastest inference time while achieving state-of-the-art performance across the official evaluation metrics of HICO-DET.}
\label{tab:hico}
  \centering
  \scriptsize
  \begin{tabular}{l c c c c c c c c}
    \toprule
    \multicolumn{2}{c}{} & \multicolumn{3}{c}{\textbf{Default}} & \multicolumn{3}{c}{\textbf{Known Object}} \\
    \cmidrule(r){3-5}\cmidrule(r){6-8}
    Method & Ext src & Full & Rare & Non Rare & Full & Rare & Non Rare & \textit{t(ms)}\\ \midrule\hline
    \multicolumn{5}{l}{\textit{Models with \textbf{external features}}} \\ \hline
    \multicolumn{1}{l|}{Verb Embedding~\cite{xu2019learning}} & \multicolumn{1}{c|}{~\cite{pennington2014glove},\cite{lu2016visual}} & {\color{gray}14.70} & {\color{gray}13.26} & {\color{gray}15.13} & {\color{gray}-} & {\color{gray}-} & {\color{gray}-} & \\
    \multicolumn{1}{l|}{TIN (R$\text{P}_{\text{D}}\text{C}_{\text{D}}$)~\cite{li2019transferable}} & \multicolumn{1}{c|}{~\cite{fang2017rmpe,li2019crowdpose}} & {\color{gray}17.03} & {\color{gray}13.42} & {\color{gray}18.11} & {\color{gray}19.17} & {\color{gray}15.51} & {\color{gray}20.26} & \\
    \multicolumn{1}{l|}{Functional Gen.~\cite{bansal2020detecting}} & \multicolumn{1}{c|}{~\cite{mikolov2013distributed}} & {\color{gray}21.96} & {\color{gray}16.43} & {\color{gray}23.62} & {\color{gray}-} & {\color{gray}-} & {\color{gray}-} & \\
    \multicolumn{1}{l|}{RPNN~\cite{zhou2019relation}} & \multicolumn{1}{c|}{~\cite{he2017mask}} & {\color{gray}17.35} & {\color{gray}12.78} & {\color{gray}18.71} & {\color{gray}-} & {\color{gray}-} & {\color{gray}-} & \\
    \multicolumn{1}{l|}{PMFNet~\cite{wan2019pose}} & \multicolumn{1}{c|}{~\cite{chen2018cascaded}} & {\color{gray}17.46} & {\color{gray}15.65} & {\color{gray}18.00} & {\color{gray}20.34} & {\color{gray}17.47} & {\color{gray}21.20} & \\
    \multicolumn{1}{l|}{No-Frills HOI~\cite{gupta2019no}$\dagger$} & \multicolumn{1}{c|}{~\cite{cao2017realtime}} & {\color{gray}17.18} & {\color{gray}12.17} & {\color{gray}18.68} & {\color{gray}-} & {\color{gray}-} & {\color{gray}-} & \\
    \multicolumn{1}{l|}{Analogies~\cite{peyre2019detecting}} & \multicolumn{1}{c|}{~\cite{mikolov2013distributed}} & {\color{gray}19.4} & {\color{gray}14.6} & {\color{gray}20.9} & {\color{gray}-} & {\color{gray}-} & {\color{gray}-} & \\ \midrule\hline
    \rowcolor[gray]{0.85}\multicolumn{9}{l}{\textit{\textbf{Models with original comparison}}} \\ \hline
    \multicolumn{1}{l|}{VSRL~\cite{gupta2015visual}} & \multicolumn{1}{c|}{\xmark} & 9.09 & 7.02 & 9.71 & - & - & - & - \\
    \multicolumn{1}{l|}{HO-RCNN~\cite{kolesnikov2019detecting}} & \multicolumn{1}{c|}{\xmark} & 7.81 & 5.37 & 8.54 & 10.41 & 8.94 & 10.85 & - \\
    \multicolumn{1}{l|}{InteractNet~\cite{gkioxari2018detecting}} & \multicolumn{1}{c|}{\xmark} & 9.94 & 7.16 & 10.77 & - & - & - & 55 \\
    \multicolumn{1}{l|}{GPNN~\cite{qi2018learning}$\dagger$} & \multicolumn{1}{c|}{\xmark} & 13.11 & 9.41 & 14.23 & - & - & - & 40\\
    \multicolumn{1}{l|}{iCAN~\cite{gao2018ican}} & \multicolumn{1}{c|}{\xmark} & 14.84 & 10.45 & 16.15 & 16.26 & 11.33 & 17.73 & 75 \\
    \multicolumn{1}{l|}{TIN (R$\text{C}_{\text{D}}$)~\cite{li2019transferable}} & \multicolumn{1}{c|}{\xmark} & 13.75 & 10.12 & 15.45 & 15.34 & 10.98 & 17.02 & 70 \\
    \multicolumn{1}{l|}{DCA~\cite{wang2019deep}} & \multicolumn{1}{c|}{\xmark} & 16.25 & 11.16 & 17.75 & 17.73 & 12.78 & 19.21 & 130  \\ \midrule
    \multicolumn{1}{l|}{\textit{\textbf{Ours}}} & \multicolumn{1}{c|}{\xmark} & \textbf{17.58} & \textbf{11.72} & \textbf{19.33} & \textbf{19.76} & \textbf{14.68} & \textbf{21.27} & \textbf{9.06} \\
    \bottomrule
  \end{tabular}
\end{table}
\subsubsection{Performance Analysis.}
We quantitatively evaluate our model on two datasets, followed by the ablation study of our proposed methods.
We use the official evaluation code for computing the performance of both V-COCO and HICO-DET.
In V-COCO, there are two versions of evaluation but most previous works have not explicitly stated which version was used for evaluation.
We have specified the evaluation scenario if it has been referred in either the literature~\cite{zhou2019relation}, authors' code or the reproduced code.
We report our performance in both scenarios for a fair comparison with heterogeneous baselines.
In both scenarios, our model outperforms state-of-the-art methods~\cite{wang2019deep}. Further, our model 
shows competitive performance compared to the baselines~\cite{xu2019learning,li2019transferable,zhou2019relation} that leverage heavy external features such as linguistic priors~\cite{pennington2014glove,mikolov2013distributed} or human pose features~\cite{fang2017rmpe,li2019crowdpose,cao2017realtime,chen2018cascaded}.
On HICO-DET, our model achieves state-of-the-art performance for both the official `Default' setting and `Known Object' setting.
For a more comprehensive evaluation of HOI detectors, we also provide the performance of recent works leverage external knowledge ~\cite{gupta2019no,wan2019pose,zhou2019relation,peyre2019detecting,xu2019learning,li2019transferable,xu2019learning,gupta2017aligned}, although the \textbf{\textit{models with External Knowledge}} are beyond the scope of this paper. 
Note that our main focus is to build a fast single-stage HOI detector from visual features.

Our \textbf{ablation study} in Table \ref{tab:ablation} shows that each component (foreground focal loss, target object classification loss $\mathcal{L}_{ij}^{cls}(\Breve{\theta})$, union matching function $\mu_u$) in our approach improves the overall performance of HOI detection.
\setlength{\tabcolsep}{12pt}
\begin{table}
\caption{Ablation Study on V-COCO test set of our model. The first row shows the performance with only the Instance Branch. It can be observed that our proposed Union Branch plays a significant role in HOI detection. The second$\sim$fourth row each shows the performance without the Target Object Classification loss $\mathcal{L}^{cls}_{ij}(\Breve{\theta})$, $\mu_u$ substituted with standard IoU score in Eq.\ref{eq:hoiscore}, Foreground Focal Loss replaced with ordinary Focal Loss, respectively.}
\label{tab:ablation}
  \centering
  \small
  \begin{tabular}{ c c c c c c }
    \toprule
    & \multicolumn{3}{c}{\textbf{UnionDet components}} & & \\
    \cmidrule(r){2-4}
    Union Branch & $FFL$ & $\mathcal{L}^{cls}_{ij}(\Breve{\theta})$ & $\mu_u$ & $Sce.\#1$ & $Sce.\#2$ \\
    \midrule
    - & - & - & - & 38.4 & 51.0 \\
    \checkmark & \checkmark & - & \checkmark & 44.8 & 53.5 \\
    \checkmark & \checkmark & \checkmark & - & 45.0 & 53.6 \\
    \checkmark & - & \checkmark & \checkmark & 46.9 & 55.6 \\
    \checkmark & \checkmark & \checkmark & \checkmark & \textbf{47.5} & \textbf{56.2} \\
    \bottomrule
  \end{tabular}
\end{table}

\subsubsection{Interaction Prediction Time.}
We measured inference time on a single Nvidia GTX1080Ti GPU. Our model achieved the fastest `end-to-end' inference time (\textbf{77.6 ms}).
However, the end-to-end inference time is not suitable for fair comparison since the end-to-end computation time of one approach may largely vary depending on the base networks or the backbone object detector.
Therefore, we here compare the \textit{\textbf{additional}} time for interaction prediction, excluding the time for object detection. The detailed analysis of end-to-end time will also be provided in the supplement.
Our approach increases the minimal inference time on top of a standard object detector by eliminating the additional pair-wise neural network inference on detected object pairs, which is commonly required in previous works.
Table~\ref{tab:vcoco} compares the inference time of the HOI interaction prediction excluding the time of the object detection.
Note that compared to other multi-stage pipelines that have heavy network structures after the object detection phase, our model additionally requires significantly less time 9.06 ms (11.7\%) compared to the base object detector.
Our approach achieves \textbf{4X$\sim$14X speed-up} compared to the baseline HOI detection models which require $40ms \sim 130ms$ per image after the object detection phase.
Since most multi-stage pipelines have extra overhead for switching heavy models between different stages and saving/loading intermediate results.
In a real-world application on a single GPU, the gain from our approach is much bigger.
\section{Conclusions}
\label{sec:conclusion}
In this paper, we present a novel one-stage human-object interaction detector.
By performing action classification and union region detection in parallel with object detection,
we achieved the \textit{fastest} inference time while maintaining comparable performance with state-of-the-art methods. Also, our architecture is generally compatible with existing one-stage object detectors and end-to-end trainable. 
Our model enables a unified HOI detection that performs object detection and human-object interaction prediction at near real-time frame rates.
Compared to heavy multi-stage HOI detectors, our model does not need to switch models across different stages and save/load intermediate results. In the real-world scenario, our model will more beneficial.



\section*{Acknowledgement}
This work was supported by the National Research Council of Science \& Technology (NST) grant by the Korea government (MSIT)(No.CAP-18-03-ETRI), National Research Foundation of Korea (NRF-2017M3C4A7065887), and Samsung Electronics, Co. Ltd.

\clearpage
%
%
\bibliographystyle{splncs04}
\bibliography{eccv2020}

\end{document}